\documentclass[sigconf]{acmart}

\usepackage{enumitem}
\usepackage{balance}
\AtBeginDocument{%
  }

\setcopyright{none}
\renewcommand\footnotetextcopyrightpermission[1]{}

\begin{document}

\title[ADMUS]{ADMUS: A Progressive Question Answering Framework Adaptable to Multiple Knowledge Sources}

\settopmatter{authorsperrow=4}

\author{Yirui Zhan}
\authornote{This work was done during the internship of Yirui Zhan at Peking University.}
\authornotemark[2]
\orcid{0009-0009-6856-958X}
\affiliation{%
  \institution{Sichuan University}
  \city{Chengdu}
  \state{Sichuan}
  \country{China}
  \postcode{43017-6221}
}
\email{zhanyirui@stu.scu.edu.cn}

\author{Yanzeng Li}
\authornote{Both authors contributed equally to this work.}
\affiliation{%
  \institution{Peking University}
  \city{Beijing}
  \country{China}
}
\email{liyanzeng@stu.pku.edu.cn}

\author{Minhao Zhang}
\affiliation{%
  \institution{Peking University}
  \city{Beijing}
  \country{China}
}
\email{zhangminhao@pku.edu.cn}

\author{Lei Zou}
\authornote{Corresponding Author.}
\affiliation{%
  \institution{Peking University}
  \city{Beijing}
  \country{China}
}
\email{zoulei@pku.edu.cn}

\renewcommand{\shortauthors}{Zhan et al.}

\begin{abstract}
With the introduction of deep learning models, semantic parsing-based knowledge base question answering (KBQA) systems have achieved high performance in handling complex questions. 
However, most existing approaches primarily focus on enhancing the model's effectiveness on individual benchmark datasets, disregarding the high costs of adapting the system to disparate datasets in real-world scenarios (e.g., multi-tenant platform).
%
Therefore, we present ADMUS, a progressive knowledge base question answering framework designed to accommodate a wide variety of datasets, including multiple languages, diverse backbone knowledge bases, and disparate question answering datasets. 
%
To accomplish the purpose, we decouple the architecture of conventional KBQA systems and propose this dataset-independent framework. Our framework supports the seamless integration of new datasets with minimal effort, only requiring creating a dataset-related micro-service at a negligible cost.
To enhance the usability of ADUMS, we design a progressive framework consisting of three stages, ranges from executing exact queries, generating approximate queries and retrieving open-domain knowledge referring from large language models. 
An online demonstration of ADUMS is available at: \url{https://answer.gstore.cn/pc/index.html}
\end{abstract}




\maketitle
\pagestyle{plain}

\section{Introduction}

With the rapid advancement of Knowledge Base (KB) research and its applications~\cite{hogan2021knowledge, wang2017knowledge}, Knowledge Base Question Answering (KBQA) has emerged as a highly popular and well-established application of KB. A KBQA system aims to retrieve or query the correct answers from the KB based on a given Natural Language Question (NLQ)~\cite{unger2014introduction, Lan2021ASO}.
The Semantic Parsing (SP) framework, recognized as a reliable solution for KBQA, which is to convert NLQ into logical structures to generate KB queries and retrieves results through graph database interfaces (e.g., SPARQL and Cypher)~\cite{hu2017answering, gu2022knowledge}.
SP-based KBQA possesses key characteristics including explainability, traceability, and explicit symbolic reasoning. Additionally, it demonstrates the capability to handle complex and multi-hop questions effectively. Therefore, it has been widely adopted by many state-of-the-art KBQA systems recently~\cite{omar2023universal, hu2021edg, kapanipathi-etal-2021-leveraging, zhang-etal-2022-crake}.
%
%

However, the existing KBQA systems predominantly concentrate on improving performance metrics assessed on benchmark datasets, disregarding the significant costs involved in re-training, re-deploying and adapting the system to different KBs and datasets (e.g., multi-tenant senario).
To overcome the limitations, we propose a flexible KBQA framework explicitly designed for seamless \underline{AD}aptation to \underline{MU}ltiple knowledge \underline{S}ources, named ADMUS.
ADMUS has the capability to accommodate multi-tenant\footnote{The multi-tenant KBQA platform is architected to support multiple tenants, allowing them to have their own isolated KBQA services, and each tenant can deploy its own dedicated KB and QA strategy without building KBQA system from the scratch.} with various user-provided language-agnostic KBs and QA datasets.
To the best of our knowledge, ADMUS is the first implemented multi-tenant KBQA system that facilitates the effortless integration of new datasets without the need of massive re-training and re-deployment.
\begin{figure*}[ht]
    \includegraphics[width=\textwidth]{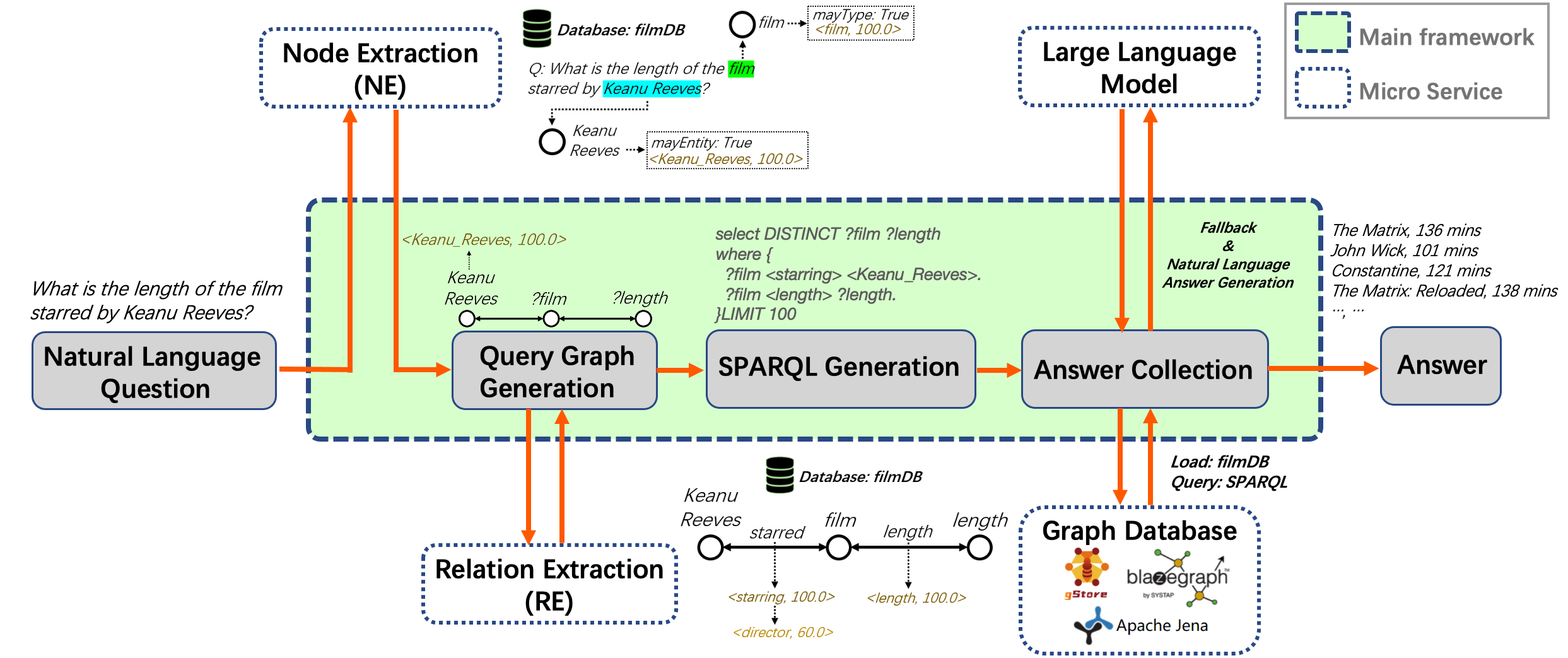}
    \caption{Architecture of ADMUS.}
    \label{fig:architecture}
\end{figure*}

Specifically, ADMUS conducts research on conventional SP-based KBQA systems and decomposes their workflow into two parts: dataset-related components, which involve node extraction and relation extraction, and dataset-independent backbone, which includes steps like semantic parsing and query graph construction.
To accomplish this decoupling, ADMUS adopts the microservice architecture, which allows for adaptive routing between different microservice providers for the various dataset-related components.
Moreover, taking into account the trade-off between the reliability of KBQA results and the usability of user experience, we develop a 3-stage progressive framework on top of the proposed ADMUS.
In the process of constructing the query graph, our framework progressively builds three types of queries: exact queries, approximate queries, and prompt-based queries that sent to a Large Language Model (LLM) fallback.
This 3-stage approach enables us to progressively provide users with precise answers in the first stage, reliable answers that may lack precision in the second stage, and responsive answers that may not be entirely reliable in the third stage{\interfootnotelinepenalty=10000\footnote{The phenomenon of hallucination~\cite{ji2023survey} may lead to misleading information for users.
Thus, in our proposed trustworthy KBQA system, the LLM serves as a fallback option for open-domain questions, ensuring a positive user experience. }}. This ensures a satisfactory user experience while balancing the reliability of the KBQA results.

The main contributions of this demonstration can be summarized as follows:
\begin{itemize}[leftmargin=*,align=left]
\item Our proposed KBQA framework, ADUMS, introduces a novel approach by decoupling dataset-related services and dataset-independent modules within the SP-based KBQA pipeline. This design enables seamless adaptation to diverse underlying KBs and lays a potential foundation for implementing the multi-tenant KBQA platform.
\item We present a progressive framework on the top of ADMUS that begins with generating exact queries and gradually transitions to approximate queries, culminating in the utilization of LLMs to ensure high usability and user satisfaction.
\item To showcase the capabilities of our framework, we implement a web demonstration that incorporates several KBs. This demonstration serves to illustrate the aforementioned abilities and provides a visual representation of all the SP-based KBQA processes.
\end{itemize}



\section{Architecture}

As shown in Figure \ref{fig:architecture}, the framework of ADMUS comprises two parts: dataset-related microservices and a dataset-independent backbone. Given a NLQ after selecting the dataset, ADMUS will connect to the correlated Node Extraction~(NE) service by dynamic routing to detect all entity mentions within the question.
These entity mentions are treated as nodes within the query graph and used as input of the Query Graph Generation~(QGG) module. Leveraging the semantic structure (e.g., Dependency Tree, Abstract Meaning Representation (AMR), etc.) of the query sentence, the QGG module constructs a semantic query graph based on the inputed entity mentions. An edge will be added between two nodes if they are in the same simple path (i.e., no other node interrupts their direct connection).
Simultaneously, the edges present in the query graph are utilized by the Relation Extraction~(RE) services to extract the relations within entity pairs. Given the size and complexity of the KBs, the RE services employ various database-related methods (e.g., predicate dictionary~\cite{xue2020value}, deep learning-based method~\cite{cord}, etc.) to identify all pairs of related nodes and their associated predicates.
At this stage, the complete query graph has been generated, and a subgraph matching strategy is employed to match the RDF graph within the database. This process generates candidate SPARQL queries\footnote{\url{http://www.w3.org/TR/rdf-sparql-query/}}. Finally, within the Answer Collection~(AC) module, the generated SPARQL queries are executed in the corresponding KB, with the results ordered based on their respective scores.
If none of the previously executed SPARQL queries yield a valid result, AC module will invoke an external LLM with prompted query as a fallback.

In this section, we will provide a detailed description of each component in ADMUS.

\subsection{Node Extraction}

The Node Extraction (NE) module, which is the entry component of ADMUS, responsible for identifying nodes within the query graph based on the input NLQ. There are four types of nodes: \textit{Entity}, \textit{Type}, \textit{Literal}, and \textit{Variable}. An \textit{Entity} represents a concrete object in the real world, while a \textit{Type} signifies the conceptual classification of entities and is commonly employed in KBs to denote entities belonging to the same class. A \textit{Literal} typically denotes the fundamental data type or literal values, which commonly serve as attribute value of entities. A \textit{Variable} is a placeholder node that may be an entity, a type or a literal.
For example, in the question illustrated in Figure \ref{fig:architecture}, ``length'' and ``Keanu Reeves'' are \textit{Entities}, and ``film'' is a \textit{Type}. Additionally, the word "what" serves as a \textit{Variable} that refers to the length of films.

For entities and types, the NE module aims to extract their mentions from the input question. These mentions are then mapped to a set of entity (type) names in the knowledge graph using dataset-specific entity linking methods~\cite{shen2014entity}.
The linking results, which include candidate entities (types) and a similarity score, are passed to the next component.

In practice, the selection of entity linking method relies on the scale and complexity of the KBs.
In the case of relatively small KBs, where all entities can be loaded into memory cache, linking entities can be achieved through exact or substring matching or neural-linker approaches~\cite{sevgili2022neural}. For large KBs like DBpedia, alternative strategies such as offline dictionary-based linker (e.g., \citet{hu2017answering}) or the utilization of third-party entity linking services~(e.g., DBpedia Lookup tool\footnote{\url{https://github.com/dbpedia/dbpedia-lookup}}) are commonly adopted. 

\subsection{Query Graph Generation}
The Query Graph Generation (QGG) module, which is an abstracted semantic parsing step from the conventional SP-based KBQA systems~\cite{hu2017answering,lin2021deep}, plays a crucial role in constructing the sketch of the query graph. This module begins with the skeleton parsed from the semantic structure (e.g., Dependency Tree) of NLQ, and arranges nodes which are extracted from the NE service.
The process of building the query graph involves determining whether there is an edge connecting every pair of nodes. Commencing with a special variable node (namely ``target node''), we employ a depth-first search algorithm to identify which nodes it is connected to. 
This is done by recursively traversing the whole semantic structure around the target node until other nodes belonging to the query graph are encountered. These traversed nodes are then assumed to be connected to the target node. 
Additionally, the newly discovered nodes are added to a queue, and the same process is iteratively executed for each node in the queue until the queue becomes empty.

At this stage, the structure of the query graph is built. Note that the resulting query graph is unstrict and allows for discrepancies with the graph pattern in the target KB. The final query graph still requires invoking the RE service to align the edges of the query graph with the predicates preset in the KB. 

\subsection{Relation Extraction}
The objective of the Relation Extraction (RE) service is to identify the predicates that connect each pair of nodes in the query graph. Given the query graph as input, the dataset-related RE service will generate the relation (aligning with predicates in KB) for node pairs along with their corresponding scores.

Similar to the NE service, the selection of the relation extraction method depends on the characteristics of the target KB. In the case of simple-structured KBs with a small number of predicates, a straightforward strategy can be employed. The predicate between two nodes is determined by identifying the common ancestor node in the semantic structure (e.g., Dependency Tree) of those two nodes.
However, for more complex KBs like DBpedia, relying solely on the above-mentioned simple strategy may lead to significant errors. Therefore, we modify the predicate detection strategy by searching within the shortest path between two nodes and considering their neighbor nodes.
To further enhance the precision of predicate detection, additional methods can be introduced. For instance, \citet{xue2020value} constructed a predicate dictionary, which can map natural language phrases to a set of possible predicates instead of exact matching.
%

\subsection{SPARQL Generation}
After processing by NE, QGG, and RE, we have obtained a coarse-grained matched superset of query graph, which contains necessary query structure and nodes.
ADUMS will process such query graph by approximate matching and searching subgraph over the target KB, generate the expected SPARQL query interactively. This approach follows a similar paradigm as previous works~\cite{hu2017answering,das2022knowledge}.
Specifically, given a query graph, we can efficiently enumerate all nodes and then consider the predicates that connect each pair of nodes. 
By constructing a subgraph from the query graph, we can employ the following subgraph matching algorithm to search for corresponding subgraphs within the target KB:

\begin{itemize}[leftmargin=*,align=left]
\item An entity node in the query graph corresponds to an entity node in the target KB.
\item A type node in the query graph corresponds to a type node in the target KB or an entity node that belongs to that type.
\item A variable node is treated as a wildcard, allowing it to potentially map to any node in the target KB.
\end{itemize}

Once $k$ subgraphs have been successfully matched, $k$ SPARQL queries are generated.
%

\subsection{Answer Collection}

Traditionally, the generated SPARQL queries are executed directly in the database engine (DB) to obtain the answers, depending totally on the executed results of DB. If a generated SPARQL query fails to exactly hit the required entity, the query will fail, and the KBQA system will return an empty response.

\begin{figure}[h]
    \centering 
    \includegraphics[width=0.85\columnwidth]{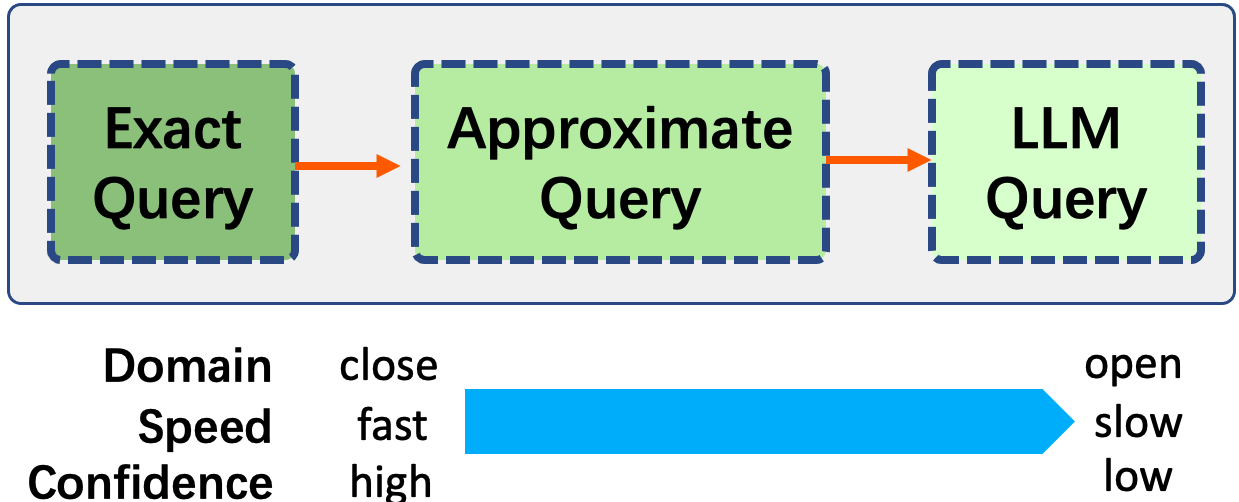}
    \caption{Three-stage Progressive Framework}
    \Description{}
    \label{fig:framework}
\end{figure}

To address the issue of user-unfriendly empty responses, ADMUS proposes a three-stage progressive framework within the Answer Collection (AC) module, as depicted in Figure 2. These three stages are as follows:
(1) \textit{Exact Query}: In this stage, if the generated SPARQL query can be executed exactly and retrieve the desired result, a trustable answer is obtained.
(2) \textit{Approximate Query}: If the generated SPARQL fails to query exactly, the AC module will rewrite the SPARQL query by transforming the exact query into an approximate query, e.g. including the ``\texttt{FILTER CONTAINS}'' statement, to expand the query's error tolerance and improve the recall of the DB execution. 
so as to obtain a reliable answer that may not be exact but still traceable and explicable.
(3) \textit{LLM Query}: If the aforementioned approximate query also fails to retrieve the required result, it suggests that the input query may be out-of-domain or cannot be resolved in SP-based manner. 
In this case, the AC module will write a LLM-involved query based on prompt templates, then remotely invoke the external LLM service to leverage its capabilities in open-domain QA or conversation. 
This fallback approach would return a LLM's answer that is objective, pertinent, and friendly, but not necessarily correct.

\section{Demonstration}

\begin{figure}
    \includegraphics[width=\linewidth]{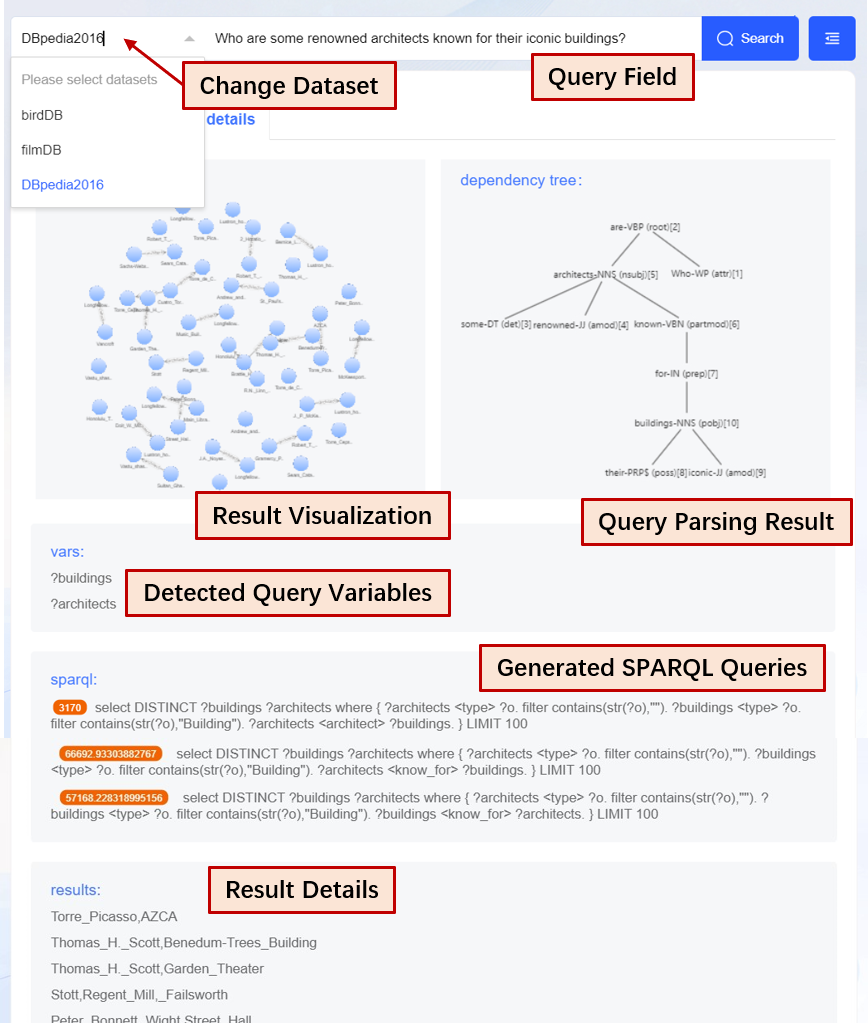}
    \centering
    \caption{Screenshot of ADMUS.}
    \label{fig:screenshot}
\end{figure}

We have utilized three datasets of varying scales and languages to demonstrate the capabilities of ADMUS, as presented in Table \ref{tab:datasets}.
To showcase the SP-based KBQA system functionality with multiple KB sources that can be easily switched, we have developed a web interface for ADMUS, as illustrated in Figure \ref{fig:screenshot}. This interface provides a comprehensive visualization of the entire process of ADMUS. It also offers the functionality to switch between different KB sources and utilize the LLM fallback feature.
With this demo, we present a potential solution to the significant challenge of the costly deployment of a new database in a KBQA system, which is crucial for a multi-tenant platform.
This web demo is publicly online available at \url{https://answer.gstore.cn/pc/index.html}.


\begin{table}[h]
    \caption{Datasets for Demonstration.}
    \centering
    \resizebox{\columnwidth}{!}{
    \begin{tabular}{c|c|c|c|c}
        \hline
        Name & Language & Triples & Entities & Predicates \\
        \hline
        birdDB & Chinese & 17,607 & 10,704 & 14 \\
        filmDB & Chinese  & 4,531,096 & 437,986 & 32 \\
        DBpedia2016 & English & 198,969,616 & 8,465,000 & 59,486 \\
        \hline
    \end{tabular}
    }
    \label{tab:datasets}
\end{table}


\section{Discussion}

Compared to end-to-end KBQA models, ADUMS can support a new dataset by adding NE and RE services rather than completely retraining a model. 
When a user wants to integrate a new NE service, there are various methods available, such as utilizing entity mention dictionaries~\cite{hu2017answering}, neural network models for Named Entity Recognition (NER)~\cite{li2020survey}, or few-shot NER approaches~\cite{huang2020few}. Similarly, when a user incorporates a new RE service, methods like predicate dictionaries~\cite{xue2020value}, Relation Detection (RD)~\cite{yu2017improved, cord}, or some of few-shot models~\cite{wen2021enhanced} can be employed. 
Regardless of the approach taken to implement a QA strategy for users' custom KB, the cost for users is negligible compared to annotating mass data, retraining model and developing an end-to-end KBQA system.

Recently, there has been a surge in popularity of QA systems and chatbots based on generative LLMs. 
In comparison to LLM-based systems, both ADMUS and traditional KBQA systems are limited to answering questions within the specific domain defined by the target KB, and lack the ability for open-domain conversations. However, LLMs often struggle with hallucination phenomena in specific domain QA~\cite{ji2023survey}, especially the factoid questions, and the answers generated by LLMs are not always convincing or reliable.
In contrast, our proposed progressive framework that allows for exact answers through SP-based KBQA while leveraging the open-domain capabilities of LLMs as a fallback. 
Furthermore, ADMUS can utilize prompt templates to convey its intermediate products of reasoning to LLMs, optimizing the final response through the Chain-of-Thought mechanism~\cite{wei2022chain}.
Additionally, LLMs and ADMUS are complementary rather than conflicting. Within various components of ADMUS, LLMs can be used to enhance the capabilities. For example, by using LLMs with prompts, it is possible to achieve few-shot NER and RD models~\cite{wei2023zero,wang2023gptner}, thereby strengthening the NE and RE modules in ADMUS and implementing a few-shot KBQA system. Additionally, ADUMS can serve as a plugin in LLM manners (e.g., LangChain~\cite{Chase_LangChain_2022}), enabling it to function as a personalized KBQA server to answer domain-specific questions exactly.

\section{Conclusion}

In summary, we propose ADMUS, a flexible and adaptable SP-based KBQA system to support various datasets in a microservices architecture. By adding new NE and RE services, ADMUS is capable of accommodating new datasets without a complete retraining or redeployment of the entire system. Moreover, the proposed three-stage progressive framework ensures both usability and credibility by integrating LLMs as a part of our system. 

In the future, we plan to leverage more advanced syntactic analysis models to enhance the generalizability of the query graph generation. We will also build more diverse datasets and more advanced NE and RE services to further improve overall system performance.

\bibliographystyle{ACM-Reference-Format}
\balance
\bibliography{BibTeX}

\end{document}